# An experimental study for early diagnosing Parkinson's disease using machine learning

Md. Taufiqul Haque Khan Tusar [a,1,*], Md. Touhidul Islam [b,2,*], Abul Hasnat Sakil [a,3]

[a] City University, Dhaka 1216, Bangladesh
[b] East West University, Dhaka 1212, Bangladesh
[1] taufiqkhantusar@gmail.com*; [2] touhid000cse@gmail.com*; [3] abulhasnatsakil.cu@gmail.com
* corresponding author

ABSTRACT

One of the most catastrophic neurological disorders worldwide is Parkinson's Disease. Along with it, the treatment is complicated and abundantly expensive. The only effective action to control the progression is diagnosing it in the early stage. However, this is challenging because early detection necessitates a large and complex clinical study. This experimental work used Machine Learning techniques to automate the early detection of Parkinson's Disease from clinical characteristics, voice features and motor examination. In this study, we develop ML models utilizing a public dataset of 130 individuals, 30 of whom are untreated Parkinson's Disease patients, 50 of whom are Rapid Eye Movement Sleep Behaviour Disorder patients who are at a greater risk of contracting Parkinson's Disease, and 50 of whom are Healthy Controls. We use MinMax Scaler to rescale the data points, Local Outlier Factor to remove outliers, and SMOTE to balance existing class frequency. Afterwards, apply a number of Machine Learning techniques. We implement the approaches in such a way that data leaking and overfitting are not possible. Finally, obtained 100% accuracy in classifying PD and RBD patients, as well as 92% accuracy in classifying PD and HC individuals.

## 1. Introduction

Parkinson's disease (PD) is defined by the abnormal degeneration of Substantia Nigra's nerve cells [1], and in contrast to somatic cells, neurons lack the inherent ability to undergo regeneration, rendering them susceptible to irreversible loss with advancing age. Hence, neuronal attrition is a phenomenon that is essentially non-compensable in nature [2]. This neurological ailment results in the permanent death of brain cells and is typically diagnosed in people over the age of 60 [3]. In terms of prevalence, morbidity, and mortality, the second-most-progressive neurodegenerative dysfunction and the most increasing neurological ailment is PD [4]. It is affecting close to 2% of adults over 65 and its rate is rising fast as the population ages [5]. Currently, over ten million people are affected by the disease globally [6]. There is now no effective tool to diagnose PD, and no neuroprotective medicines can prevent or delay its progression [7], which is why PD has become a highly serious medical and societal concern [5]. Early detection is crucial to minimizing the consequences of PD, which may extend life expectancy and assure medication effectiveness. However, the diagnosis of PD typically requires a variety of experiments, invasive procedures, and empirical diagnostic tests. However because of diverse equipment setups, these methods are neither cost-effective nor feasible. Furthermore, professionals may lead to poor decisions owing to excessive workload. Artificial intelligence-based solutions can assist specialists and physicians in automatically detecting certain illnesses, as well as reducing clinicians' workload [8].

Numerous studies have investigated the potential of deep learning and machine learning techniques to detect and diagnose PD. In [3], a parallel CNN framework with nine layers was proposed, achieving an accuracy of 86.9% and an F1-score of 91.70%. The PPMI data in [6] was utilized to develop multiple deep learning models with various machine learning techniques. The ensemble of these models achieved an accuracy of 96.68% and an F1-score of 97.58%. The authors in [9] developed a CNN-based Computer-Aided Diagnosis (CAD) approach using electroencephalogram signals of 10 male and 10 female PD patients, achieving 88.25%, 84.71%, and 91.77% accuracy, sensitivity, and specificity, respectively. In [10], the authors used an audio signal-based PD dataset and a novel method called CLS to detect PD patients and healthy subjects. The KNN classifier achieved the best performance with 91.5% accuracy. Additionally, the authors in [11] trained a DNN using a voice signal dataset of 42 patients, achieving 62.73% accuracy for the UPDRS score and 81.67% accuracy for the Motor UPDRS score. PCA2 was applied for dimensionality reduction in the features in [12], and the method achieved a clustering accuracy of 94.06% by ensembling the Genetic Algorithm and K-means. In [13], Random Forest, SVM, and Naive Bayes were applied to a motor function-related dataset grouped into three subsets, achieving accuracies of 96.7%, 95.00%, and 93.30% with SVM, RF, and NB, respectively, for the 2C(60) data subset and 86.7%, 91.7%, and 91.7% accuracy, respectively, with SVM, RF, and NB for the 2C(IH) data subset. In [14], the authors reviewed research from 2007 to 2019 and concluded that simple wearable MIoT devices and ML applications can achieve above 90% accuracy in classifying PD patients and healthy subjects. Finally, the authors in [15] proposed a novel and improved staging for PD that can classify different stages with 97.46% accuracy using the AdaBoost-based model. According to the authors in [16], ML classifiers are preferable options for achieving high accuracy with little computing time, while deep learning classifiers are superior when accuracy is the primary concern, although they require more resources and time. The best ensemble algorithms are AdaBoost and XGBoost, with the necessary trade-off between execution time and performance.

Our study utilized an open-source dataset comprising 130 patients classified into three groups: 50 Healthy Control (HC) subjects, 50 individuals at a high risk of developing Parkinson's disease (PD) due to Rapid Eye Movement (REM) Sleep Behavior Disorder (RBD), and 30 untreated PD patients. We applied various data pre-processing techniques and multiple learning techniques to classify these individuals, with the primary objective of developing a machine learning-based model for the early detection of PD using clinical characteristics, voice features, and motor examination. Our research addresses the pressing need for a cost-effective and feasible diagnostic tool for PD, given its increasing prevalence and lack of effective treatments.

Our study's contribution lies in the development of a model that can aid physicians and specialists in automatically detecting PD, thereby potentially reducing their workload and improving early detection rates. This research's novelty is the application of machine learning techniques to clinical characteristics, voice features, and motor examination, an area that has not been extensively explored in previous studies. The remainder of this paper is organized as follows. The methodology is thoroughly described in section 2, and the study's findings are presented in section 3. Finally, we conclude our study in section 4, discussing limitations and future directions.

## 2. Method

We conducted a multi-method analysis to detect PD, RBD, and HC. The original dataset was divided into two subsets, PDRBD and PDHC, with 30 PD & 50 RBD patients and 30 PD & 50 HC subjects in each, respectively. Both subsets were further divided into train-validation (85%) and test (25%) groups. The pre-processing techniques were trained on the train-validation set and then applied to both sets. To prepare the data for analysis, encoding was applied first, followed by rescaling using MinMax Scaler, removing outliers with Local Outlier Factor (LOF), and balancing the class ratio with Synthetic Minority Oversampling Technique (SMOTE). Statistical Machine Learning (ML) algorithms were then employed to train classifiers, and their performance was evaluated using 10-stratified 10-fold cross-validation. The best models were selected and their hyperparameters were optimized using Grid Search CV before testing on the test data. The workflow of our experiment is shown in Fig. 1, and the procedure for training and validating the classifiers is detailed in Algorithm 1 (Fig. 2). All methods were implemented using Python (3.10.4) with SK-learn (1.0.1) and other packages in Google Colaboratory.

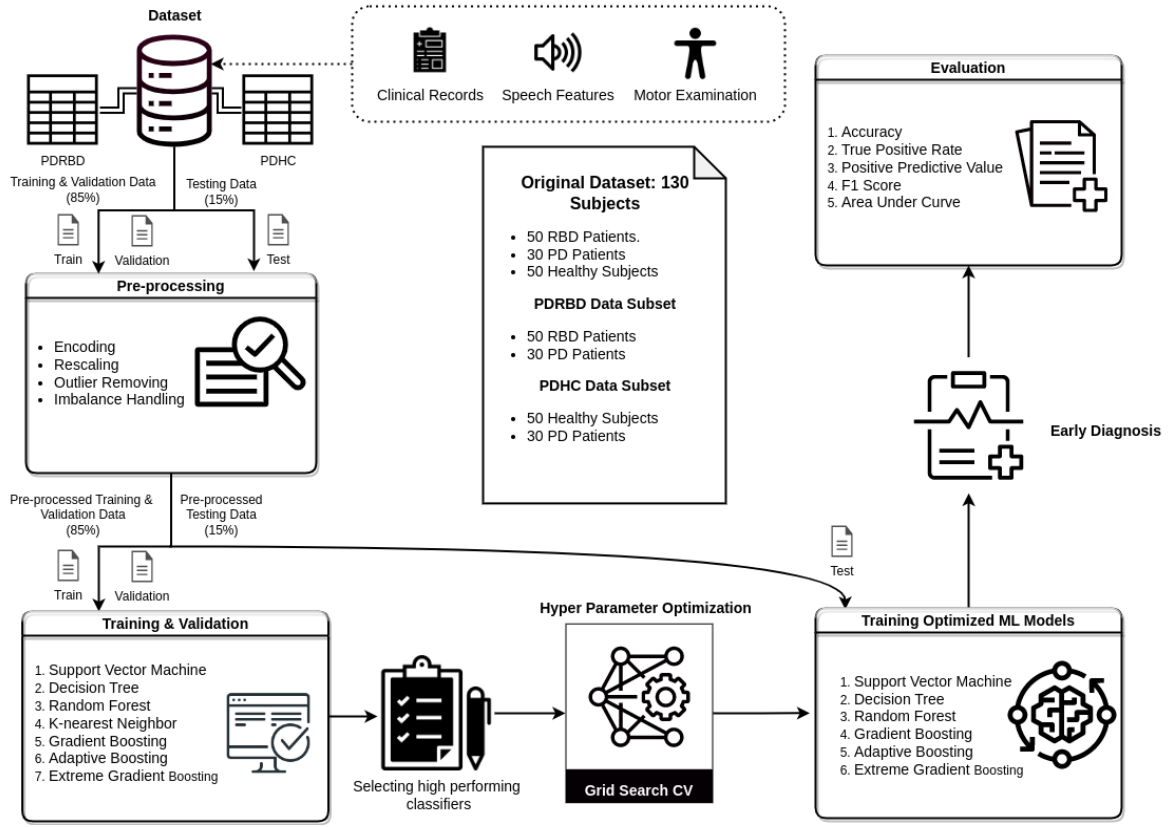

Fig. 1. The experimental method's workflow

---

**Algorithm 1.**   Training the Machine Learning classifiers

**INPUT** $x$: Training features, $y$: Training labels
**OUTPUT** Trained ML classifiers

$z \leftarrow minmax\ normalization$ technique
$lof \leftarrow Local\ Outlier\ Factor$ technique
$smote \leftarrow Synthetic\ Minority\ Oversampling\ Technique$
$cv \leftarrow 10\ stratified\ 10\ fold\ cross\ validation$
$hyper_{opt} \leftarrow hyper\ parameter\ optimization$
$m_{cv} \leftarrow models\ obtained\ by\ cv$
$m_{opt} \leftarrow optimized\ model\ by\ hyper_{opt}$

**function** $TRAINING\ (x, y)$
  **function** $PREPROCESSING\ (x)$
    $x_{norm} = z(x)$         //normalize each value of x using minmax normalization.
    $x_{lof} = lof(x_{norm})$     //remove outliers from $x_{norm}$ using the local outlier factor.
    $x_{smote} = smote(x_{lof})$   //balance the class frequency using SMOTE.
    $x_{clean} = x_{smote}$
    **return** $x_{clean}$
  **end function**
  **function** $VALIDATION\ (x_{clean}, y)$
    $m_{cv} = cv(x_{clean}, y)$   //returns models with higher cross-validation accuracy.
    **return** $m_{cv}$
  **end function**

---

```
m_opt = hyper_opt(m_cv, x_clean, y)    //optimize parameters of the selected models from m_cv
m = train m_opt             // train ml models from m_opt
return m
end function
```

Fig. 2. Algorithm of the experimental method

### 2.1. Data Set

This study utilized a dataset gathered during a medical session organized by Jan Hlavnika et al. [17] in the Czech Republic between 2014 and 2016. The dataset consisted of 130 individuals, including 50 healthy subjects (41 men, 9 women), 50 patients with RBD (41 men, 9 women), and 30 untreated PD patients (21 men, 9 women). The dataset comprised 64 features of three types, namely clinical characteristics of the examinee, speech features, and motor examination results of each examinee. To provide insight into the class distribution of the data, Fig. 3 was included, and the clinical characteristics of the examinee were displayed in Table 1. The original dataset was divided into two subsets, with the PDRBD subset remaining unchanged. The PDHC subset contained 24 features, including clinical characteristics and speech examination results, since the motor examination results were not available for healthy subjects. Attempting to simulate these results could result in inaccuracies. The data subsets were then pre-processed and subjected to machine learning techniques, as shown in Fig. 4, which depicts the original dataset and the two data subsets.

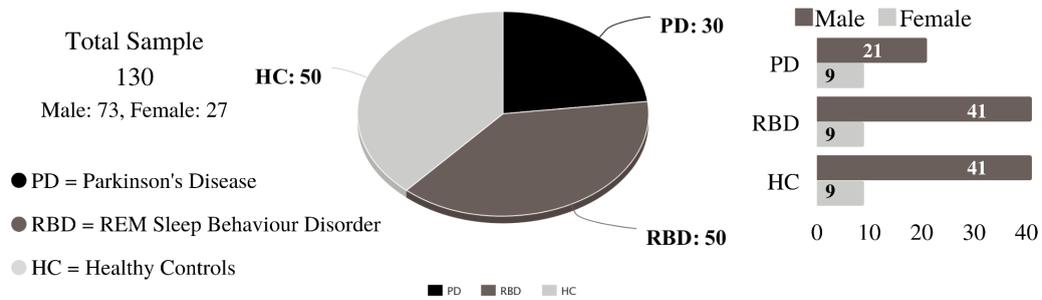

Fig. 3. Class Frequency in the Dataset

Table 1. Clinical characteristics of the examinee

| Feature | PD | RBD | HC |
|---|---|---|---|
| Gender | Men: 21<br>Women: 9 | Men: 41<br>Women: 9 | Men: 41<br>Women: 9 |
| Age | Min: 40.00<br>Avg: 63.98<br>Max: 83.00 | Min: 40.00<br>Avg: 64.92<br>Max: 83.00 | Min: 34.00<br>Avg: 64.93<br>Max: 79.00 |
| Positive histories of Parkinson's disease in family | Yes: 02<br>No: 28 | Yes: 01<br>No: 49 | NA |
| Age of disease onset (years) | Avg: 63.40 | Avg: 59.16 | NA |
| Duration of disease from first symptoms (years) | Avg: 1.63 | Avg: 5.76 | NA |
| Antidepressant therapy | Yes: 03<br>No: 27 | Yes: 07<br>No: 43 | Yes: 0<br>No: 50 |
| Antiparkinsonian medication | No: 30 | No: 50 | No: 50 |
| Antipsychotic medication | No: 30 | No: 50 | No: 50 |
| **Benzodiazepine medication** | Yes: 04<br>No: 26 | Yes: 08<br>No: 42 | Yes: 0<br>No: 50 |
| **Clonazepam (mg/day)** | Avg: 0.066<br>Max: 1.00 | Avg: 0.075<br>Max: 2.00 | Avg: 0.00<br>Max: 0.00 |

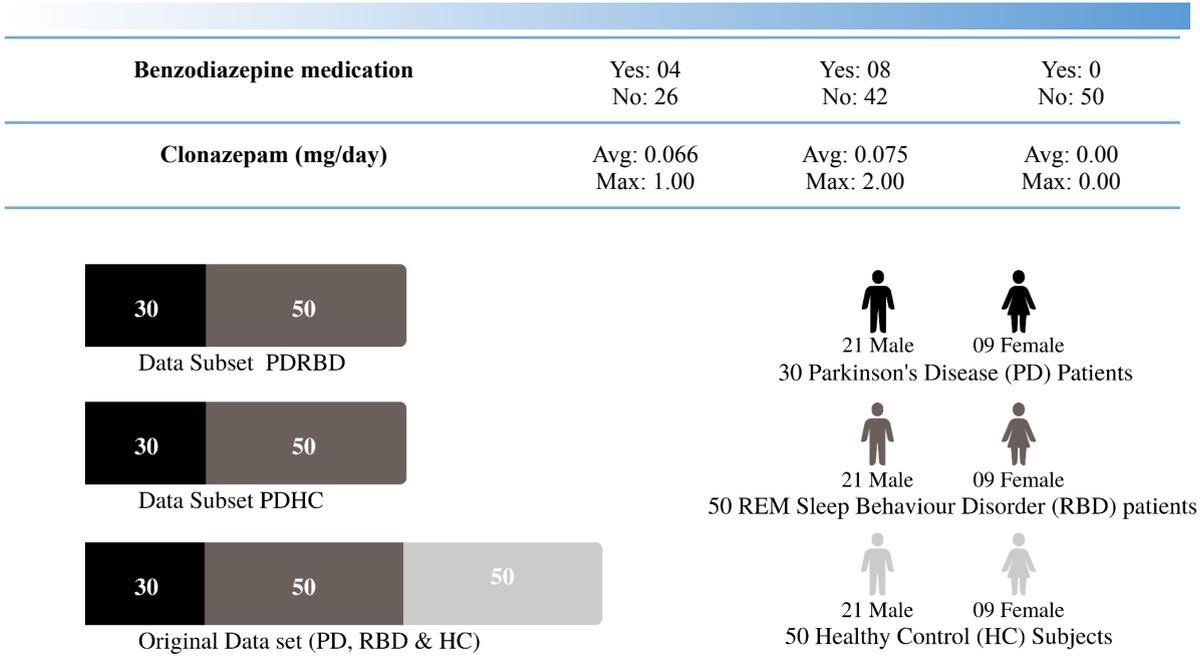

Fig. 4. Data subsets and original dataset

### 2.2. Pre-processing

*2.2.1. Encoding*

The data set includes 64 features, six of which are categorical. We have encoded these categorical features into 1's or 0's as shown in Table 2.

Table 2. Encoding

| Features | 1 | 0 |
|---|---|---|
| Gender | Female | Male |
| Positive history of Parkinson's disease in family | Yes | No |
| Antidepressant therapy | Yes | No |
| Anttiparkinsonian medication | Yes | No |
| Antipsychotic medication | Yes | No |
| Benzodiazepine medication | Yes | No |

*2.2.2. Normalization*

We employed the widely-used MinMax Scaler as the data normalization technique to rescale the numerical feature values between 0 and 1. The formulation of the MinMaxScaler can be expressed as (1), where $x$ denotes the original value, $x_{min}$ refers to the minimum value of the feature, $x_{max}$ denotes the maximum value of the feature, and $x_{norm}$ denotes the normalized value.

$$x_{norm} = \frac{(x - x_{min})}{x_{max} - x_{min}} \qquad (1)$$

The MinMax Scaler is particularly useful for handling features with a broad range of values. By rescaling the features into a uniform range, the influence of large values that may dominate the learning algorithm is removed, allowing all features to contribute equally to the learning process. Moreover, normalization is an effective way to reduce the impact of noise and outliers in the data, resulting in more robust models.

*2.2.3. Local Outlier Factor (LOF)*

Outliers have a disastrous effect on the training of an ML model. It can occur in the training of a model using redundant data. According to the authors of [18], when appropriate outlier identification approaches are used, ML techniques on healthcare data will produce more accurate results in diagnosing disorders. We applied density-based outlier detection, called LOF to eliminate outliers.

*2.2.4. Synthetic Minority Oversampling Technique (SMOTE)*

Real-world data sets across practically all domains include class imbalance, and it has an impact on accuracy. Because the models underperform when dealing with a small number of data classes [19]. The SMOTE performs well on the small imbalanced dataset to balance the class ratio. Here, Fig. 5 and 6 illustrate the class frequency of the data subsets after removing outliers and after applying SMOTE to the data subsets respectively.

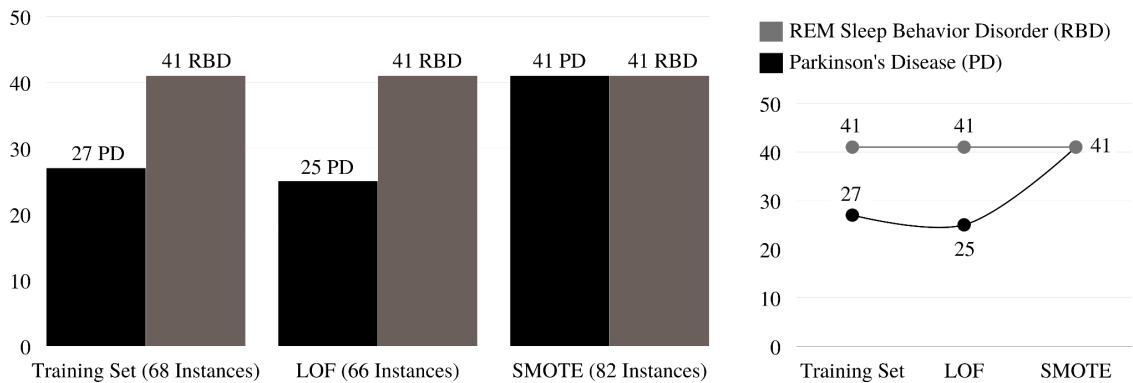

Fig. 5. Data frequency in various stages of preprocessing in the PDRBD

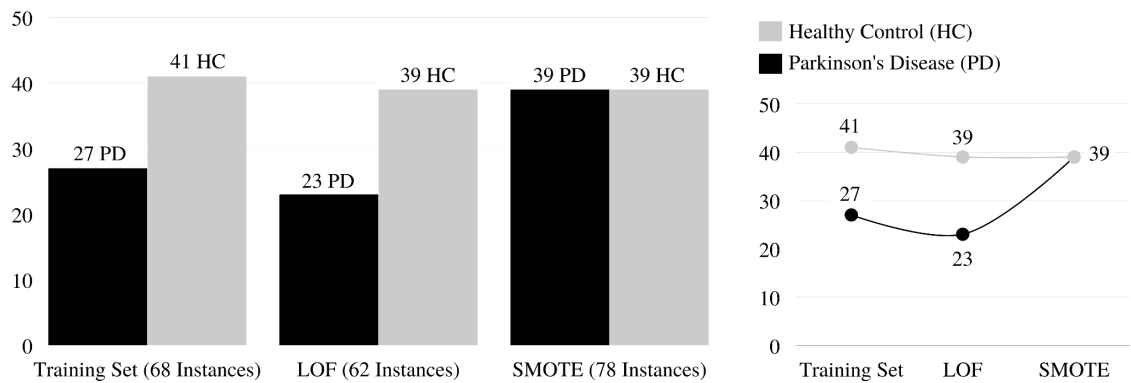

Fig. 6. Data frequency in various stages of preprocessing in the PDHC

**2.3. Training and Validation**

The data subsets named PDRBD and PDHC were separately partitioned and preprocessed prior to training and validating ML models. Eight distinct classification algorithms named SVM, XGBoost, AdaBoost, GBoost, KNN, Logistic Regression (LR), Random Forest (RF), and Decision Tree (DT) were used to train and validate the ML models for identifying PD, RBD, and HC utilizing the train-validation split. A 10-stratified 10-fold CV was applied to validate the ML models, which repeated the 10-fold CV 10 times and maintained the class ratio according to the training dataset [20]. After that, the most effective classifiers were selected for further experimentation based on the validation results. The training and validation approach is shown in Fig. 7.

Fig. 8 shows the validation accuracy of different ML models on PDRBD and PDHC. The training & validation outcome on PDRBD has been illustrated in Fig. 8(a), where SVM, RF, KNN, and LR provide validation accuracy of 97.79%, 99.01%, 87.47%, and 97.58% respectively. Also, DT, GBoost, AdaBoost, and XGBoost achieve a validation accuracy of 100%. Here, the validation losses of SVM, RF, KNN, and LR are 7.27%, 3.35%, 11.14%, and 5.15% respectively, whereas the DT, GBoost, AdaBoost, and XGBoost have 0% validation losses. Fig. 9(a) shows the accuracy and loss graph of the PD and RBD classifiers found by the validation step.

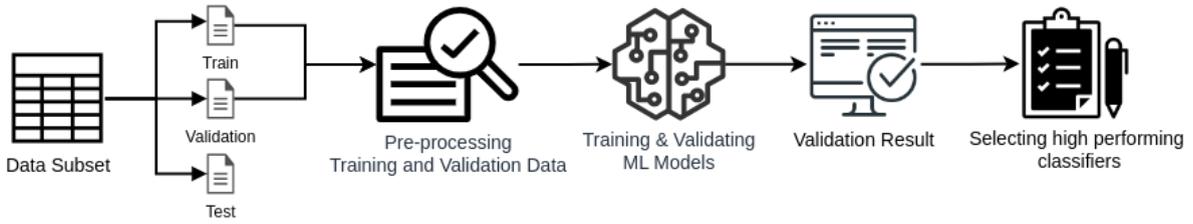

Fig. 7. Training and validation approach for each data subset

Similarly, the training & validation outcome on PDHC has been illustrated in Fig. 8(b), where SVM, DT, RF, KNN, LR, GBoost, AdaBoost, and XGBoost provide validation accuracy of 75.41%, 68.45%, 75.20%, 59.89%, 72.11%, 71.46%, 64.88%, and 70.16% respectively. Here, the validation losses of SVM, DT, RF, KNN, LR, GBoost, AdaBoost, and XGBoost are 12.94%, 16.50%, 12.75%, 15.80%, 13.52%, 14.15%, 15.24%, and 14.60% respectively, as shown in Fig. 9(b).

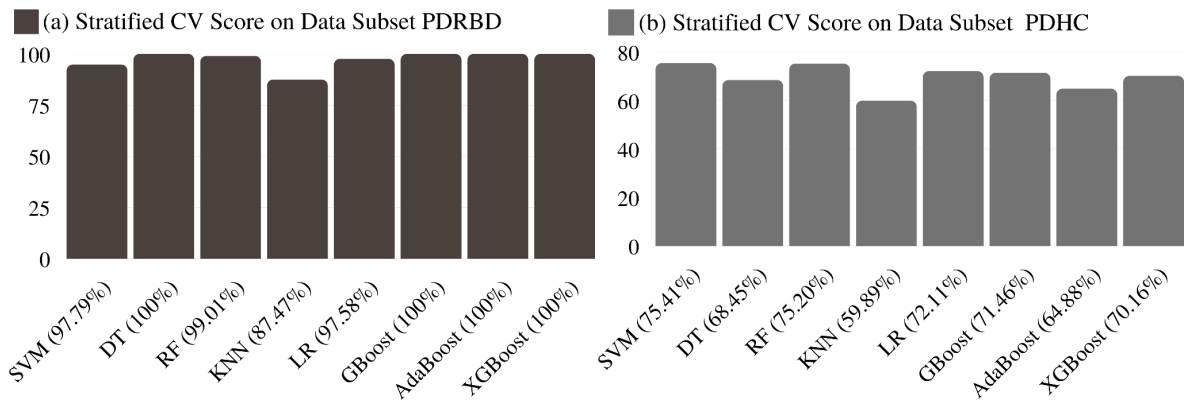

Fig. 8. Validation result on data subsets (a) PDRBD and (b) PDHC

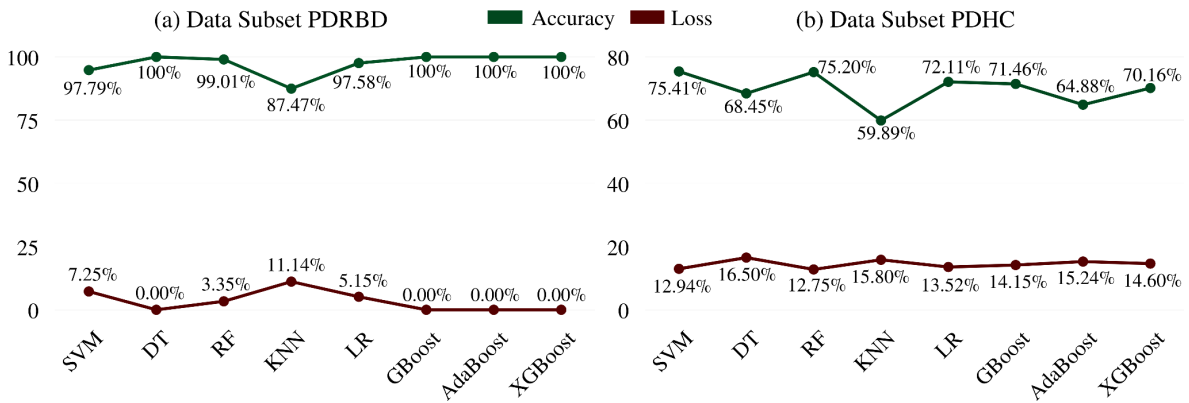

Fig. 9. Validation accuracy vs validation loss on data subsets (a) PDRBD and (b) PDHC

### 2.4. Hyper Parameter Optimization

In this section, we have selected the best-performing algorithms during validation to build final ML models to classify PD, RBD, and HC. Here we select the SVM, DT, RF, LR, GBoost, AdaBoost, and XGBoost for both data subsets. In this hyperparameter optimization stage, we used Grid Search CV for tuning the best-performing algorithm's parameters, where the number of splits is 5 and it repeats 5 times for each classifier. Grid search CV is a widely used method for hyperparameter tuning that exhaustively searches the hyperparameter space to find the optimal combination of parameters.

For the SVM, the regularization parameter is 1, the kernel is RBF, the heuristic is true, and the tolerance is 1e − 3. Gini impurity was used for information gain in the DT, and the best split was chosen for further procedure. The RF uses 1000 DT as the estimator and the Gini impurity for information gain. In the LR, the regularization parameter is 1, the solver is Limited-memory BFGS, and the tolerance is 1e − 4. For the GBoost, the learning rate is 0.01, MSE with an improvement score by Friendman has been used to ensure the split's quality, LR has been used to optimize loss, and 1000 boosting stages have been used. For AdaBoost, the learning rate is 1.0, and SAMME.R has been used as the ground estimator. Finally, for the XGBoost, 1000 GBoost has been used, and the learning rate is 0.3.

We used hyperparameter tuning to optimize the models' performance and find the optimal combination of hyperparameters that minimize the error rate. Grid Search CV was selected because it is a widely used and reliable method for hyperparameter tuning. By fine-tuning the hyperparameters, we were able to improve the models' performance and achieve the high accuracy rates reported in this study.

### 3. Results and Discussion

We collected and separated the dataset into two subsets: PDRBD and PDHC. After that pre-processed the data subsets and applied various ML algorithms. We trained and validated the ML models using 10-stratified 10-fold cross-validation and selected the high-performing algorithms for both subsets. We then optimized the hyperparameters of these algorithms using Grid Search CV and evaluated their performance using Accuracy, True Positive Rate (TPR), Positive Predictive Value (PPV), F1-score, and AUC Score respectively. Tables 3 and 4 show the evaluation outcomes of the ML models which have been calculated using the following (2) - (6).

$$Accuracy = \frac{TP + TN}{TP+TN+FP+FN} \quad (2)$$

$$TPR\ or\ Recall = \frac{TP}{TP+FN} \quad (3)$$

$$PPV\ or\ Precision = \frac{TP}{TP+FN} \quad (4)$$

$$F1 - score = \frac{2*PPV*TPR}{PPV+TPR} \quad (5)$$

$$AUC\ (f) = \frac{\Sigma t_0 \epsilon D^0 * \Sigma t_0 \epsilon D^1 * 1[f(t_0)<f(t_1)]}{|D^0|*|D^1|} \quad (6)$$

Here, TP, FP, TN, FN are True Positive, False Positive, True Negative and False Negative respectively. $1[f(t_0) < f(t_1)] = 1$, while $f(t_0) < f(t_1)$ is True otherwise 0. $D^0$ is a set of negative examples and $D^1$ is a set of positive examples.

Table 3 displays the results of all ML models used for classifying the PD and RBD, indicating 100% accuracy, PPV, TPR, F1-score, and AUC. However, Table 4 shows that the accuracy for the PDHC data subset ranges from 0.67 to 0.92, and the highest AUC score is 0.89 for the decision tree (DT) model. The discrepancy in performance can be attributed to the characteristics of each subset. The PDRBD dataset consists of PD and RBD cases that share similarities in symptoms and clinical features, and all 64 attributes were utilized to classify them. In contrast, the PDHC dataset includes PD and healthy controls, and only 24 features were utilized due to the absence of motor examination results for the healthy subjects. As a result, it is more challenging to differentiate between these two groups. Further analysis shows that the AdaBoost and GBoost algorithms perform well in terms of accuracy (92%), PPV (92% and 94% respectively), TPR (92%) and F1-score (92%) for PDHC, indicating their effectiveness in classifying PD and healthy controls with limited features.

However, the analysis is limited due to the unavailability of significant features for PDHC, such as a comprehensive motor examination overview. These features are crucial for distinguishing PD and healthy controls. The lack of these features may have contributed to the model's difficulty in correctly classifying PD and HC. Nonetheless, this limitation is due to the unavailability of data, and conducting a more comprehensive study with a complete dataset is necessary to achieve better performance. In Table 5, a comparative analysis with recent studies indicates that the approaches described in this paper perform better in detecting PD patients.

Table 3. Evaluation outcome from the data subset PDRBD

| Algorithm | Accuracy | PPV | TPR | F1 - score | AUC |
|---|---|---|---|---|---|
| SVM | **1.00** | 1.00 | 1.00 | 1.00 | **1.00** |
| DT | **1.00** | 1.00 | 1.00 | 1.00 | **1.00** |
| RF | **1.00** | 1.00 | 1.00 | 1.00 | **1.00** |
| LR | **1.00** | 1.00 | 1.00 | 1.00 | **1.00** |
| GBoost | **1.00** | 1.00 | 1.00 | 1.00 | **1.00** |
| AdaBoost | **1.00** | 1.00 | 1.00 | 1.00 | **1.00** |
| XGBoost | **1.00** | 1.00 | 1.00 | 1.00 | **1.00** |

Table 4. Evaluation outcome from the data subset PDHC

| Algorithm | Accuracy | PPV | TPR | F1 - score | AUC |
|---|---|---|---|---|---|
| SVM | 0.75 | 0.72 | 0.75 | 0.73 | 0.78 |
| DT | 0.83 | 0.83 | 0.83 | 0.83 | **0.89** |
| RF | 0.67 | 0.67 | 0.67 | 0.67 | 0.85 |
| LR | 0.83 | 0.90 | 0.83 | 0.84 | 0.85 |
| GBoost | **0.92** | **0.94** | **0.92** | **0.92** | 0.81 |
| AdaBoost | **0.92** | **0.92** | **0.92** | **0.92** | 0.81 |
| XGBoost | 0.75 | 0.88 | 0.77 | 0.77 | 0.83 |

Table 5. A comparative study with related works

| No | Best Model and Year | Accuracy |
|---|---|---|
| 1. | CNN (2019) [3] | 86.90 % |
| 2. | CNN (2020) [9] | 88.25 % |
| 3. | KNN (2022) [10] | 91.50 % |
| 4. | SVM (2020) [21] | 91.25 % |
| 5. | ANN (Levenberg– Marquardt algorithm) (2021) [22] | 95.89 % |
| 6. | KNN (2019) [23] | 94.55 % |
| 7. | Random Forest (feature selection: Genetic Algorithm) (2022) [24] | 95.58 % |
| 8. | XGBoost (2019) [25] | 95.39 % |
| 9. | Random Forest (2022) [26] | 97.00 % |
| 10. | Light Gradient Boosting (2020) [27] | 84.10 % |
| 11. | SVM (2018) [28] | 92.00 % |
| 12. | Bat (2020) [29] | 96.74 % |
| 13. | Bi-LSTM (2021) [30] | 84.29 % |
| | **Performance of this study** | |
| | [Performance on PDRBD] | |
| 14. | SVM | 100.00 % |
| | DT | 100.00 % |
| | RF | 100.00 % |
| | LR | 100.00 % |
| | GBoost | 100.00 % |
| | AdaBoost | 100.00 % |
| | XGBoost | 100.00 % |
| | [Performance on PDHC] | |
| | GBoost | 92.00 % |
| | AdaBoost | 92.00 % |

## 4. Conclusion

In this study, we investigated the detection of Parkinson's Disease (PD), Rapid Eye Movement Sleep Behavior Disorder (RBD), and Healthy Control (HC) subjects using a complex and imbalanced dataset. Through the application of diverse data pre-processing techniques and multiple ML algorithms, we achieved unprecedented accuracy of 100% in detecting early-stage PD and RBD patients using SVM, DT, RF, LR, GBoost, AdaBoost, and XGBoost classifiers. Our GBoost and AdaBoost classifiers also demonstrated a commendable accuracy of 92% in discerning PD and HC subjects. However, limitations of the study include the relatively small dataset used, limited domain knowledge, and lack of real-time data. Incorporating larger and real-time datasets, domain knowledge for outlier detection and attribute importance analysis can enhance model efficacy. Our study highlights the potential of ML in assisting specialists and physicians in early PD detection, leading to reduced clinical workload and improved detection rates. Future research can refine the models by integrating larger datasets and domain knowledge to develop a cost-effective and feasible diagnostic tool for PD. Our promising results pave the way for continued exploration in this field, and we encourage further research to advance the understanding and detection of PD and related neurological disorders.

**References**

[1] C. Blauwendraat, M. A. Nalls, and A. B. Singleton, "The genetic architecture of Parkinson's disease," Lancet Neurol., vol. 19, no. 2, pp. 170–178, 2020, doi: 10.1016/S1474-4422(19)30287-X.

[2] M. Dhanawat, D. K. Mehta, S. Gupta, and R. Das, "Understanding the pathogenesis involved in Parkinson's disease and potential therapeutic treatment strategies," Cent. Nerv. Syst. Agents Med. Chem., vol. 20, no. 2, pp. 88–102, 2020, doi: 10.2174/1871524920666200705222842.

[3] H. Gunduz, "Deep learning-based Parkinson's disease classification using vocal feature sets," IEEE Access, vol. 7, pp. 115540–115551, 2019, doi: 10.1109/access.2019.2936564.

[4] GBD 2016 Parkinson's Disease Collaborators, "Global, regional, and national burden of Parkinson's disease, 1990-2016: a systematic analysis for the Global Burden of Disease Study 2016," Lancet Neurol., vol. 17, no. 11, pp. 939–953, 2018, doi: 10.1016/S1474-4422(18)30295-3.

[5] B. Brakedal, L. Toker, K. Haugarvoll, and C. Tzoulis, "A nationwide study of the incidence, prevalence and mortality of Parkinson's disease in the Norwegian population," NPJ Parkinsons Dis., vol. 8, no. 1, p. 19, 2022, doi: 10.1038/s41531-022-00280-4.

[6] W. Wang, J. Lee, F. Harrou, and Y. Sun, "Early detection of Parkinson's disease using deep learning and machine learning," IEEE Access, vol. 8, pp. 147635–147646, 2020, doi: 10.1109/access.2020.3016062.

[7] A. Govindu and S. Palwe, "Early detection of Parkinson's disease using machine learning," Procedia Comput. Sci., vol. 218, pp. 249–261, 2023, doi: 10.1016/j.procs.2023.01.007.

[8] A. S. Albahri et al., "A systematic review of trustworthy and explainable artificial intelligence in healthcare: Assessment of quality, bias risk, and data fusion," Inf. Fusion, vol. 96, pp. 156–191, 2023, doi: 10.1016/j.inffus.2023.03.008.

[9] S. L. Oh et al., "A deep learning approach for Parkin- son's disease diagnosis from EEG signals," Neural Comput. Appl, vol. 32, pp. 10927–10933, 2020, doi: 10.1007/s00521-018-3689-5.

[10] F. Demir, K. Siddique, M. Alswaitti, K. Demir, and A. Sengur, "A simple and effective approach based on a multi-level feature selection for automated Parkinson's disease detection," J. Pers. Med., vol. 12, no. 1, p. 55, 2022, doi: 10.3390/jpm12010055.

[11] S. Grover, S. Bhartia, and A. Yadav, Seeja K.R., " Predicting severity of Parkinson's disease using deep learning," Procedia Comput. Sci, vol. 132, pp. 1788–1794, 2018, doi: 10.1016/j.procs.2018.05.154.

[12] M. T. Islam, S. R. Rafa, and M. G. Kibria, "Early prediction of heart disease using PCA and hybrid genetic algorithm with k-means," in 2020 23rd International Conference on Computer and Information Technology (ICCIT), 2020, doi: 10.1109/ICCIT51783.2020.9392655.

[13] F. Cavallo, A. Moschetti, D. Esposito, C. Maremmani, and E. Rovini, "Upper limb motor pre-clinical assessment in Parkinson's disease using machine learning," Parkinsonism Relat. Disord., vol. 63, pp. 111–116, 2019, doi: 10.1016/j.parkreldis.2019.02.028.

[14] M. Belić, V. Bobić, M. Badža, N. Šolaja, M. Đurić-Jovičić, and V. S. Kostić, "Artificial intelligence for assisting diagnostics and assessment of Parkinson's disease-A review, " Clin," Clin. Neurol. Neurosurg, vol. 184, no. 105442, 2019, doi: 10.1016/j.clineuro.2019.105442.

[15] R. Prashanth and S. Dutta Roy, "Novel and improved stage estimation in Parkinson's disease using clinical scales and machine learning," Neurocomputing, vol. 305, pp. 78–103, 2018, doi: 10.1016/j.neucom.2018.04.049.

[16] J. Goyal, P. Khandnor, and T. C. Aseri, "A Comparative Analysis of Machine Learning classifiers for Dysphonia-based classification of Parkinson's Disease," Int. J. Data Sci. Anal., vol. 11, no. 1,

pp. 69–83, 2021, doi: 10.1007/s41060-020-00234-0.

[17] J. Hlavnička, R. Čmejla, T. Tykalová, K. Šonka, and E. Růžička, "Automated analysis of connected speech reveals early biomarkers of Parkinson's disease in patients with rapid eye movement sleep behaviour disorder," Sci. Rep, vol. 7, 2017, doi: 10.1038/s41598-017-00047-5.

[18] A. Miti, "A critical overview of outlier detection methods," Comput. Sci. Rev, vol. 38, 2020, doi: 10.1016/j.cosrev.2020.100306.

[19] H. Kaur, H. S. Pannu, and A. K. Malhi, "A systematic review on imbalanced data challenges in machine learning: Applications and solutions," ACM Comput. Surv, vol. 52, pp. 1–36, 2020, doi: 10.1145/3343440.

[20] M. T. Haque Khan Tusar, M. T. Islam, and F. I. Raju, "Detecting chronic kidney disease(CKD) at the initial stage: A novel hybrid feature-selection method and robust data preparation pipeline for different ML techniques," in 2022 5th International Conference on Computing and Informatics (ICCI), 2022, doi: 10.1109/ICCI54321.2022.9756094.

[21] O. Yaman, F. Ertam, and T. Tuncer, "Automated Parkinson's disease recognition based on statistical pooling method using acoustic features," Med. Hypotheses, vol. 135, no. 109483, p. 109483, 2020, doi: 10.1016/j.mehy.2019.109483.

[22] G. Pahuja and T. N. Nagabhushan, "A comparative study of existing machine learning approaches for Parkinson's disease detection," IETE J. Res., vol. 67, no. 1, pp. 4–14, 2021, doi: 10.1080/03772063.2018.1531730.

[23] J. S. Almeida et al., "Detecting Parkinson's disease with sustained phonation and speech signals using machine learning techniques," Pattern Recognit. Lett., vol. 125, pp. 55–62, 2019, doi: 10.1016/j.patrec.2019.04.005.

[24] R. Lamba, T. Gulati, H. F. Alharbi, and A. Jain, "A hybrid system for Parkinson's disease diagnosis using machine learning techniques," Int. J. Speech Technol., vol. 25, no. 3, pp. 583–593, 2022, doi: 10.1007/s10772-021-09837-9.

[25] I. Nissar, D. Rizvi, S. Masood, and A. Mir, "Voice-based detection of Parkinson's disease through ensemble machine learning approach: A performance study," EAI Endorsed Trans. Pervasive Health Technol., vol. 5, no. 19, p. 162806, 2019, doi: 10.4108/eai.13-7-2018.162806.

[26] I. Ahmed, S. Aljahdali, M. Shakeel Khan, and S. Kaddoura, "Classification of Parkinson disease based on patient's voice signal using machine learning," Intell. autom. soft comput., vol. 32, no. 2, pp. 705–722, 2022, doi: 10.32604/iasc.2022.022037.

[27] I. Karabayir, S. M. Goldman, S. Pappu, and O. Akbilgic, "Gradient boosting for Parkinson's disease diagnosis from voice recordings," BMC Med. Inform. Decis. Mak., vol. 20, no. 1, p. 228, 2020, doi: 10.1186/s12911-020-01250-7.

[28] S. Lahmiri, D. A. Dawson, and A. Shmuel, "Performance of machine learning methods in diagnosing Parkinson's disease based on dysphonia measures," Biomed. Eng. Lett., vol. 8, no. 1, pp. 29–39, 2018, doi: 10.1007/s13534-017-0051-2.

[29] R. Olivares et al., "An optimized brain-based algorithm for classifying Parkinson's disease," Appl. Sci. (Basel), vol. 10, no. 5, p. 1827, 2020, doi: 10.3390/app10051827.

[30] C. Quan, K. Ren, and Z. Luo, "A deep learning based method for Parkinson's disease detection using dynamic features of speech," IEEE Access, vol. 9, pp. 10239–10252, 2021, doi: 10.1109/access.2021.3051432.